\documentclass[Afour,sageh,times]{sagej}

\usepackage{moreverb,url}

\usepackage[colorlinks,bookmarksopen,bookmarksnumbered,citecolor=red,urlcolor=red]{hyperref}

\newcommand\BibTeX{{\rmfamily B\kern-.05em \textsc{i\kern-.025em b}\kern-.08em
T\kern-.1667em\lower.7ex\hbox{E}\kern-.125emX}}

\def\volumeyear{2016}

\usepackage{xspace}
\usepackage{subcaption}

\newcommand{\SO}[1]{\mathbb{SO}(#1)}
\newcommand{\SE}[1]{\mathbb{SE}(#1)}
\newcommand{\R}[1]{\mathbb{R}^#1}
\newcommand{\RplusSO}[1]{\mathbb{R}^#1\text{ and }\mathbb{SO}(#1)}
\newcommand*{\eg}{e.g.\@\xspace}

\newcommand*{\ie}{i.e.\@\xspace}
\newcommand{\energy}[1]{\|#1\|^2}

\graphicspath{{./figures/}{./}}

\begin{document}

\runninghead{Ovr\'en and Forss\'en}

\title{Trajectory Representation and 
Landmark Projection for 
%Rolling Shutter
Continuous-Time Structure from Motion}

\author{Hannes Ovr\'en\affilnum{1} and Per-Erik Forss\'en\affilnum{1}}

\affiliation{\affilnum{1}Link\"oping University, Sweden}

\corrauth{Hannes Ovr\'en,
Computer Vision Laboratoty,
Department of Electrical Engineering,
Link\"oping University,
SE-581 83 Link\"oping, Sweden}

\email{hannes.ovren@liu.se}

\begin{abstract}
This paper revisits the problem of continuous-time structure from motion,
and introduces a number of extensions that improve convergence and efficiency.
The formulation with a $\mathcal{C}^2$-continuous spline for the trajectory
naturally incorporates inertial measurements, 
as derivatives of the sought trajectory.
We analyse the behaviour of split interpolation on $\SO3$ and on $\R3$, and a joint interpolation on $\SE3$,
and show that the latter implicitly couples the direction of translation and rotation.
Such an assumption can make good sense for a camera mounted on a robot arm, but not for hand-held
or body-mounted cameras.
Our experiments show that split interpolation on $\RplusSO3$ is preferable over $\SE3$ interpolation in all tested cases.
Finally, we investigate the problem of landmark reprojection on rolling shutter cameras,
and show that the tested reprojection methods give similar quality, 
while their computational load varies by a factor of 2.
\end{abstract}


\maketitle
\section{Introduction}
Structure from motion on video, is a variant of the
Simultaneous Localisation And Mapping (SLAM) problem, which by now is one of
the classical problems in robotics \citep{bailey06}. Structure from motion on video has a 
wide range of applications, such as 3D mapping \citep{engel16}, video
stabilization \citep{kopf14}, and autonomous navigation 
\citep{bailey06}. Traditionally such systems used discrete-time
camera poses, while this paper considers the more recent continuous-time formulation
\citep{furgale15}. Many SLAM systems exploit a combination of sensors
for robustness; LIDAR, cameras, and inertial sensors (typically
gyroscopes and accelerometers) are popular choices. It is well known
that cameras and inertial sensors are complementary, and thus useful
to combine. Primarily this is because inertial measurements have
biases, that can be estimated during fusion with camera
measurements. In addition, cameras often provide very accurate relative
pose, but not absolute scale, and camera-only structure from motion fails in the absence
of scene structure.

\begin{figure}[t]
  \begin{center}
    \includegraphics[width=\columnwidth]{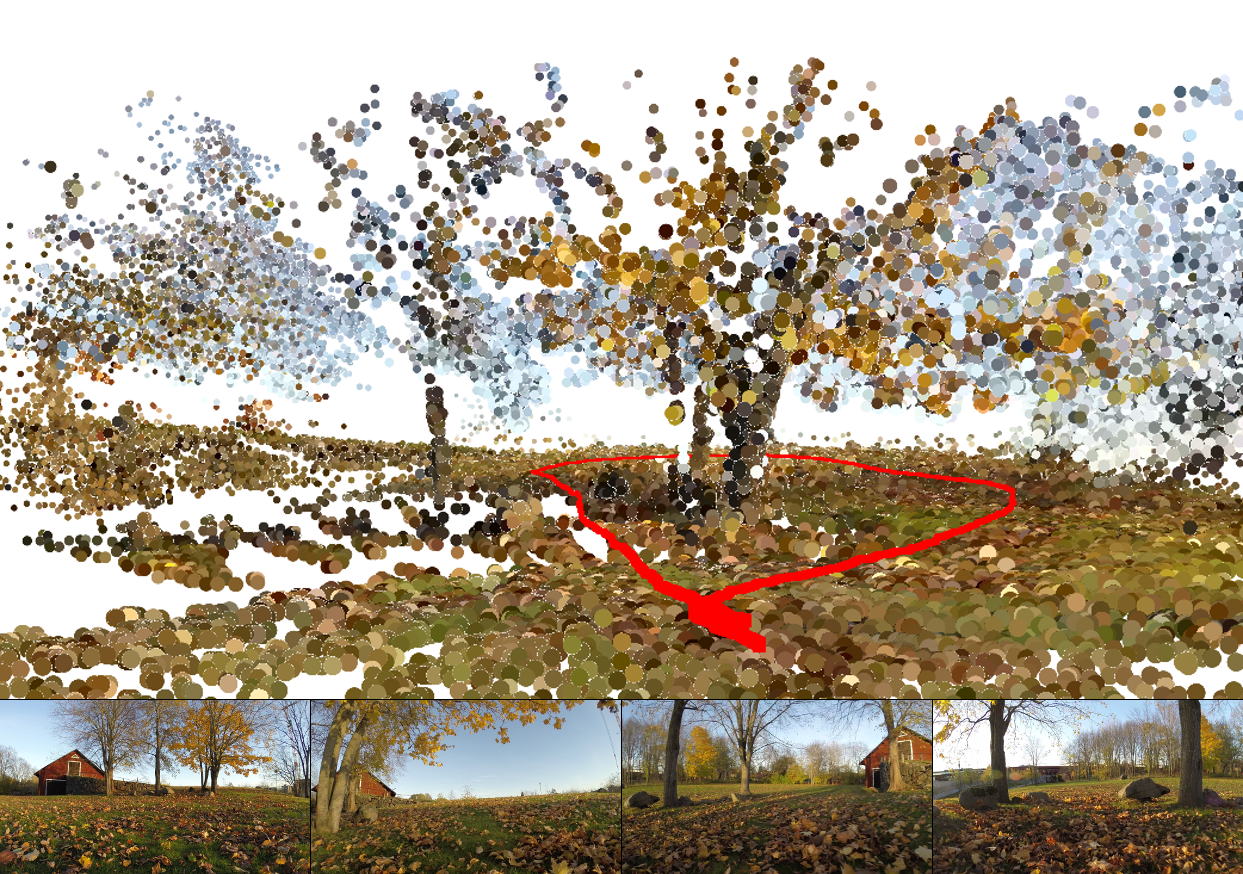}
    \caption{Rendered model estimated on the {\bf RC-Car} dataset, using split interpolation. 
    Top: model rendered using Meshlab. Bottom: Sample frames from dataset.}
    \label{fig:example_rccar}
  \end{center}
\end{figure}

Platforms that house both cameras and inertial sensors are now very
common.
Examples include most current smartphones and tablets, but also some action cameras,
\eg newer models from GoPro.
Nearly all such platforms use cameras with an electronic {\it rolling shutter}
mechanism, that acquires each frame in a row-by-row fashion. This
lends itself naturally to continuous-time motion models,
as the camera has a slightly different pose in each image row.

Classical structure from motion treats camera trajectories as a set
of discrete poses \citep{triggs00}, but by replacing the poses with spline knots,
 we obtain the continuous-time formulation, which is
used on rolling shutter cameras for video structure from motion \citep{hedborg12}.
A useful property of the continuous pose representation, introduced by \cite{furgale12}, 
is that its derivatives can predict measurements from an inertial measurement unit (IMU), which simplifies fusion of data from cameras and IMUs, and multi-sensor
platforms in general \citep{furgale15}. Continuous-time structure from motion is also crucial in
camera-IMU calibration when the camera has a rolling shutter
\citep{ovren15,furgale15,lovegrove13}. Compared to classical structure from motion,
the continuous-time version has a moderate increase in complexity, due to reduced
sparsity of the system Jacobian as shown by \cite{hedborg12}.

\subsection{Contributions}

In this paper we revisit the continuous-time structure from motion problem with inertial measurements, and
rethink several design choices:
\begin{itemize}
\item We replace the $\SE3$-based interpolation used in
  the \emph{Spline Fusion} method \citep{lovegrove13,patron-perez15} with a split
  interpolation in $\RplusSO3$. This leads to a trajectory representation
  that does not couple rotation and translation in a screw motion,
  see \figurename~\ref{fig:interaction}, and is better suited to \eg, hand-held camera motions.
\item We compare the split and $\SE3$ trajectory representations
  theoretically, and in a series of both synthetic and real data experiments.
\item We compare the performance
and efficiency of three previously proposed ways to incorporate reprojection
time into the optimization \citep{hedborg12,furgale12, lovegrove13, kim16}.
\item For completeness, we also describe our recently published {\it
    spline error weighting} approach to better balance the residuals
  in the optimization problem, and to automatically set the
  spline knot spacing based on desired trajectory accuracy \citep{ovren18a}.
\end{itemize}

The main goal of the paper is to help other researchers make informed
choices when designing their continuous-time structure from motion systems.

\subsection{Related work}
\label{sec:related_work}
The classical pose interpolation approach in computer animation is to
independently interpolate the camera orientation in the orientation group
$\SO3$ and the camera positions in the vector space
$\R3$ \citep{kim95}.

In robotic animation it is instead common to do 
direct interpolation on the special Euclidean
group $\SE3$ \citep{crouch99}. Recently, such a
direct interpolation on $\SE3$ was applied to the continuous-time structure from motion problem, 
by integrating the $\SE3$ spline into an optimization framework
\citep{lovegrove13,patron-perez15}. This formulation generalizes the
orientation interpolation of \cite{kim95} to
$\SE3$. Several recent continuous-time structure from motion papers use the
$\SE3$ approach \citep{patron-perez15,kerl15,kim16,anderson15}, while
others use separate interpolation of pose and orientation 
\citep{furgale15,oth13}. In the following sections, we
analyse the two approaches theoretically, and also compare them experimentally.

When re-projecting a landmark in a frame there is an additional
complication in the rolling shutter case. As one image coordinate
(typically the image row) corresponds to observation time, the reprojection
of a landmark at time $t$ will not necessarily end up at the row
corresponding to that time.
Early methods handled this by setting the reprojection time to the
landmark observation time \citep{hedborg12,furgale12}. This was
improved upon by \cite{oth13} who also linearize the reprojection time error
and convert it to a spatial error covariance.
\cite{lovegrove13} instead use the Newton method to iteratively find a
reprojection time with a consistent row coordinate, and this 
approach is also followed by \cite{kerl15}.
Yet another approach is to add a projection time parameter for each
landmark observation, as well as a cost term for the projection time deviation
\citep{kim16}. In the experiments, we refer to this approach as {\it lifting}, 
which is the common term for elimination of alternating
optimization by adding variables and constraints \citep{zach14}. No
previous publication has compared these choices, instead each paper
makes a hard commitment to one of the methods. In \cite{furgale15}
some of the choices are discussed, but a comparison is left for future work.

\subsection{Paper overview}

The remainder of the paper is organized as follows. 
In section \ref{sec:visual_inertial_fusion} we introduce the
visual-inertial fusion problem that is the context of this paper.
In section \ref{sec:projection} we describe three methods for
rolling shutter landmark projections,
and in section \ref{sec:trajectories} we present
two different choices of continuous trajectory representation.
Finally, in \ref{sec:experiments} we evaluate our methods experimentally, and
section \ref{sec:conclusions} summarizes the paper and gives an outlook.

Illustrations and plots are best viewed in colour.

\begin{figure*}[t]
\centering
\begin{subfigure}{0.49\textwidth}
  \begin{center}
    \includegraphics[width=\columnwidth]{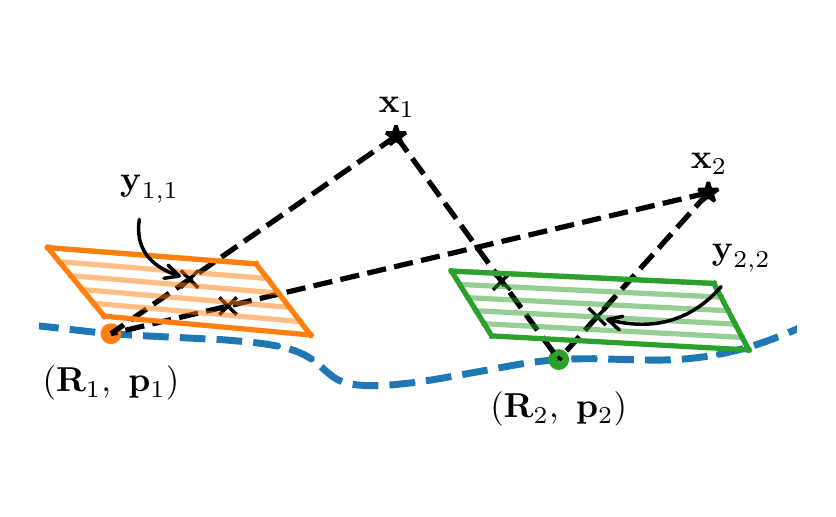}
    \caption{Global shutter projection}
    \label{fig:sfm_gs}
  \end{center}
\end{subfigure}
\begin{subfigure}{0.49\textwidth}
  \begin{center}
    \includegraphics[width=\columnwidth]{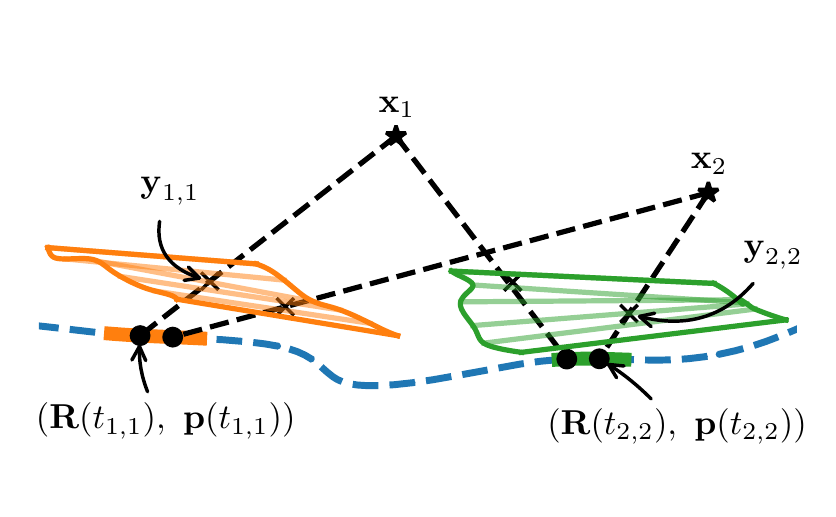}
    \caption{Rolling shutter projection}
    \label{fig:sfm_rs}
  \end{center}
\end{subfigure}
\caption{Structure from motion under global and rolling shutter geometry. 
Here, ${\bf x}_k$ is a 3D landmark which is projected to an image observation, ${\bf y}_{k, n}$, in image $n$.
Cameras are represented by their image plane, where we also show a limited number of the image rows.
On the camera trajectory (dashed, blue line) we indicate the time instance (global shutter),
or time span (rolling shutter),
when the image was captured.
}
\label{fig:sfm}
\end{figure*}

\section{Visual-inertial fusion}
\label{sec:visual_inertial_fusion}
This work is an extension of the \emph{Spline fusion}
visual-inertial fusion framework introduced by \cite{lovegrove13}.
In this section we outline how the Spline fusion method works,
and also summarize the improvements to robustness of the framework,
introduced by \cite{ovren18a}.

\subsection{Video structure from motion}
In structure from motion, the goal is to estimate 
camera poses, and 3D structure,
from a set of images.
If the images are from video, or are taken in sequence,
the camera poses can be thought of as a trajectory over time.

A camera pose consists of a rotational component ${\bf R} \in \SO3$,
and a translational component ${\bf p} \in \R3$.
In standard structure from motion, the camera path is simply
the set of all camera poses, with one pose per image, $n$:
\begin{align}
  {\bf T}_n = ({\bf R}_n, {\bf p}_n) \,.
  \label{eq:pose_discrete}
\end{align}
We follow the convention in \cite{patron-perez15}, and define the
pose such that ${\bf T}$ is a transformation from the body (\ie camera) to
the global coordinate frame.

The objective is then to find the camera poses, and 3D points
that minimize the cost function
\begin{align}
  J(\mathcal{T}, \mathcal{X}) = 
  \sum_{{\bf T}_n \in \mathcal{T}}
  \sum_{{\bf x}_k \in \mathcal{X}}
  \|
    {\bf y}_{k, n} - \pi({\bf T}^{-1}_n {\bf x}_k)
  \|^2 \,.
  \label{eq:sfm_basic}
\end{align}
Here, $\mathcal{X}$ is the set of all 3D points, $\mathcal{T}$ is the set of all camera poses,
and ${\bf y}_{k, n}$ is the observation of 3D point $k$ in image $n$.
The function $\pi(\cdot)$ projects a 3D point in the camera coordinate frame to the image plane,
using some camera model.
This formulation of the structure from motion objective is called \emph{bundle adjustment} \citep{triggs00}.
We illustrate the structure from motion geometry in Figure \ref{fig:sfm_gs}.

\subsection{Rolling shutter}
\label{sec:rolling_shutter}
In the previous section, we assumed that there is one camera pose per image, such that all
pixels are captured at the same time.
Such cameras are said to have a \emph{global shutter}.

Most cameras available today are however equipped with a \emph{rolling shutter} \citep{elgamal05}.
Here, the image is read out from the sensor one row at a time, 
\ie different rows are captured at different times.
If the camera is moving while the image is captured, then we no longer have
a single camera pose per image, but instead one camera pose per row.
We illustrate the rolling shutter geometry in Figure \ref{fig:sfm_rs},
where the camera pose at the
row that corresponds to image observation ${\bf y}_{k,n}$
is denoted $({\bf R}(t_{k,n}), {\bf p}(t_{k,n}))$.

It has been shown \citep{hedborg12} that ignoring rolling shutter when minimizing \eqref{eq:sfm_basic} reduces accuracy, and can even lead to reconstruction failures.

\subsection{Continuous-time structure from motion}
To handle the rolling shutter problem,
the standard, or discrete-time, formulation of structure from motion in \eqref{eq:pose_discrete}
can be modified to instead model the camera trajectory as a continuous-time function
\begin{align}
  {\bf T}(t) = ({\bf R}(t), {\bf p}(t)) \,.
\end{align}
Instead of being restricted to a set of discrete camera poses, we can now determine
the camera pose at any time instance $t$.
There are many ways to construct ${\bf T}(t)$, but argubly the most common approach
is to model it as some kind of spline.

Given this new representation, we modify the cost function \eqref{eq:sfm_basic} to
\begin{align}
  J(\mathcal{T}, \mathcal{X}) = 
  \sum_{{\bf y}_{k, n} \in \mathcal{Y}}
  \|
    {\bf y}_{k, n} - \pi({\bf T}^{-1}(t_{k, n}){\bf x}_k)
  \|^2 \,.
  \label{eq:ct_sfm_basic}
\end{align}
where $\mathcal{Y}$ is the set of all image observations, and $\mathcal{X}$ is still the set of 3D points.
However, $\mathcal{T}$ is no longer a set of discrete camera poses, 
but is instead the set of \emph{trajectory parameters}.
The exact nature of the trajectory parameters depends on how
we choose to model the trajectory.

With a continuous-time formulation, structure from motion can be
solved on both rolling shutter, and global shutter cameras by
minimizing the same cost function \eqref{eq:ct_sfm_basic}. 
There are however some practical aspects regarding how the landmarks are projected into the
camera, which we will further investigate in section \ref{sec:projection}. 

Next, we will show another new possibility: incorporating inertial measurements
in the bundle adjustment formulation.

\subsection{Inertial measurements}
An IMU consists of a gyroscope, which measures angular velocities, $\boldsymbol{\omega}$,
and an accelerometer, which measures linear accelerations, ${\bf a}$.
These measurements are direct observations of motion,
and are a useful addition to the trajectory estimation problem.
\cite{lovegrove13} therefore extend \eqref{eq:ct_sfm_basic} to also
include residuals for the gyroscope and accelerometer measurements:
\begin{align}
  J(\mathcal{T}, \mathcal{X}) = 
  \sum_{{\bf y}_{k, n} \in \mathcal{Y}}&
  \|
    {\bf y}_{k, n} - \pi({\bf T}^{-1}(t_{k, n}){\bf x}_k)
  \|^2 \notag
  \\
  +\sum_m&||\boldsymbol{\omega}_m-\nabla_{\omega}{\bf T}(t_m)||^2_{{\bf W}_g}
  \label{eq:ct_sfm_inertial}
  \\
  +\sum_l&||{\bf a}_l-\nabla^2_{a}{\bf T}(t_l)||^2_{{\bf W}_a}\,. \notag
\end{align}
The operators $\nabla_\omega$ and $\nabla^2_a$ represent inertial sensor models which
predict gyroscope and accelerometer values given the trajectory model ${\bf T}(t)$,
using analytic differentiation.
The norm weight matrices ${\bf W}_g$ and ${\bf W}_a$ are used to balance the
three modalities fairly.
We show how to set the norm weight matrices in section \ref{sec:sew_weights}.

For best results, the inertial sensor models, $\nabla_\omega$ and $\nabla^2_a$,
should model the used sensors as well as possible.
At the very least they should account for a constant measurement bias,
however, more advanced models that include \eg axis misalignment, or
time-varying biases, are also possible.
In section \ref{sec:trajectories} we derive basic inertial
sensor models for the trajectories which we are interested in.

Looking at the IMU residuals in \eqref{eq:ct_sfm_inertial}, we can see that there are two things that
makes it problematic to use a discrete camera trajectory here.
Firstly, the IMU measurement timestamps do not in general coincide with the frame times.
This is partly because the IMU is usually sampling at a much higher rate than the camera.
With a trajectory consisting only of discrete poses, it is not obvious how to extract 
a pose for these intermediate timestamps.
The continuous-time formulation does not have this problem, since it allows
us to determine the camera pose at any given time instance.

Secondly, the IMU residuals require us to compute derivatives of the
trajectory, to get angular velocity and linear acceleration, respectively.
With a discrete-time trajectory, these derivatives are not available.
A continuous-time trajectory can, however, be built such that the required
derivatives exist. To avoid derivatives, discrete time systems commonly
use sensor integration instead. However, whenever the sensor bias
is updated, the sensor integration has to be recomputed. Much effort
has thus been spent to improve performance of sensor integration in the
discrete pose case \citep{forster15}.

Since we need second order derivatives to compute the acceleration,
it is crucial that the trajectory representation ${\bf T}(t)$ is
$\mathcal{C}^2$-continuous.
This is the reason why cubic B-splines are a popular choice.

\subsection{Splined trajectories}
\label{sec:splines}
Splines are an excellent choice for representing a 
continuous trajectory because their derivatives can
be easily computed analytically.
To introduce the general concept of splines, we will
first describe it in only one dimension.
In section \ref{sec:trajectories} we then describe how splines
can be used to model a continuous camera pose.

A spline consists of a set of \emph{control points},
$\boldsymbol{\Theta} = (\theta_1,\ldots,\theta_K)$,
which are positioned at \emph{knots},
$(t_1,\ldots,t_K)$,
in time.
The value of the spline at a specific time $t$ is computed
from the control points, which are weighted by a \emph{basis function}, $B(t)$.
If the knots are evenly spaced, $\Delta t$ apart, we have $t_k = k\Delta t$,
and say that the spline is \emph{uniform}:

\begin{align}
  f(t|\boldsymbol{\Theta}) = \sum_{k=1}^K \theta_k B(t-k\Delta t)\,.
  \label{eq:spline_1d}
\end{align}

Fitting data to a (uniform) spline means to optimize the spline control points, 
$\boldsymbol{\Theta}$, such that the shape of the spline matches the
measurements. The knot spacing, $\Delta t$, is a hyper parameter,
and in section \ref{sec:sew_knot_spacing} we show one way to set
it to an appropriate value.

\subsection{Spline Error Weighting}
\label{sec:spline_error_weighting}
Before we attempt to minimize \eqref{eq:ct_sfm_inertial},
using a splined trajectory, there are three hyper-parameters that must be
set to apropriate values:
the knot spacing, $\Delta t$,
and the IMU residual norm matrices, ${\bf W}_a$ and ${\bf W}_g$.

\cite{lovegrove13}, who introduced \eqref{eq:ct_sfm_inertial},
used a fixed knot spacing value of $\Delta t = 0.1$,
and set the norm weight matrices to the inverse covariance of the respective measurement noises.
In \cite{ovren18a} we showed why these choices are suboptimal, and
derived a robust method to set these values.
We will now give a summary of this method, which is called \emph{Spline Error Weighting}.

\subsubsection{Selecting the IMU weights.}
\label{sec:sew_weights}
If we use inverse covariances to weight the inertial measurements in \eqref{eq:ct_sfm_inertial},
then we make the implicit assumption that the chosen trajectory parameterization can
perfectly represent the real motion.

However, a high $\Delta t$ (sparse spline) results in a smooth trajectory which might not be able to fully
represent the real motion.
In this case the residuals in \eqref{eq:ct_sfm_inertial} will consist of two error terms:
the measurement noise, and a \emph{spline approximation error}.
The Spline fusion method only accounts for the former, and in \cite{ovren18a} 
we showed that ignoring the approximation error leads to reconstruction failures.

\begin{figure}[bt]
  \begin{center}
    \includegraphics[width=\columnwidth]{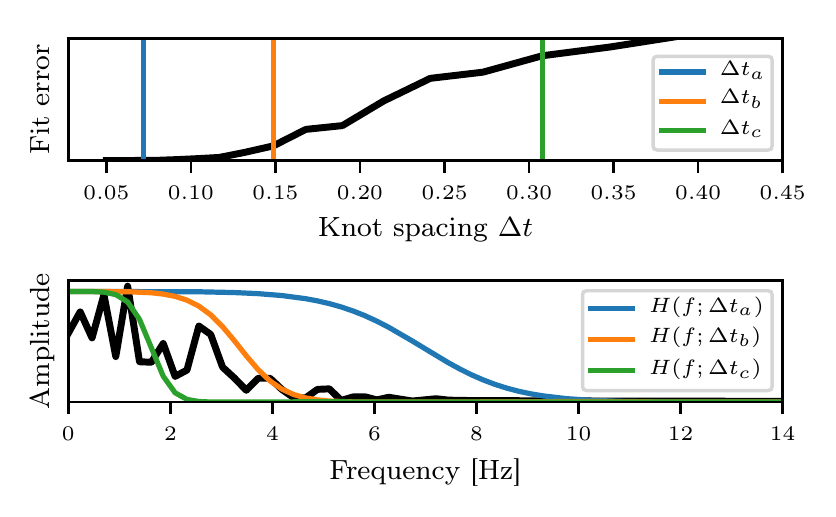}
    \caption{Top: Interpolation error for a test signal $x(t)$ as a function of spline knot spacing $\Delta t$. 
             Bottom: The frequency spectrum of $x(t)$ (black) together with the frequency function $H(f; \Delta t)$ for different choices of $\Delta t$.}
    \label{fig:spline_dt_freq}
  \end{center}
\end{figure}

Spline fitting can be characterized in terms of a frequency response
function, $H(f)$, see \cite{unser1993}. 
In this formulation, a signal $x(t)$ with the Discrete Fourier Transform (DFT)
$X(f)$ will have the frequency content $(H\cdot X)(f)$ after spline
fitting.
In Figure \ref{fig:spline_dt_freq} we show examples of how the spline interpolation function $H(f)$,
and the spline fit error,
depends on the choice of knot spacing.

By denoting the DFT of the frequency response function by the vector
${\bf H}$, and the DFT of the signal by ${\bf X}$, we can express the
error introduced by the spline fit as:
\begin{equation}
{\bf E}=(1-{\bf H})\cdot {\bf X}\,.
\label{eq:approximation_error}
\end{equation}
This results in an approximation error variance
\begin{equation}
\hat{\sigma}_e^2=\energy{{\bf E}}/N=\energy{(1-{\bf H})\cdot{\bf X}}/N\,,
\label{eq:approximation_variance}
\end{equation}
where $N$ is the number of samples.
The residual weight matrices in \eqref{eq:ct_sfm_inertial} are then computed as
\begin{align}
{\bf W}_r = \frac{1}{\hat\sigma_r^2}{\bf I}~~\text{where}~~
\hat{\sigma}_r^2 = \hat{\sigma}_e^2+\hat{\sigma}_f^2\,.
\label{eq:residual_error}
\end{align}
Here $\hat{\sigma}_f^2$ is a filtered version of the sensor noise variance $\sigma_n^2$,
to account for the fact that ${\bf X}$ used in \eqref{eq:approximation_error}
already contains this noise.

\begin{figure}[t]
\includegraphics{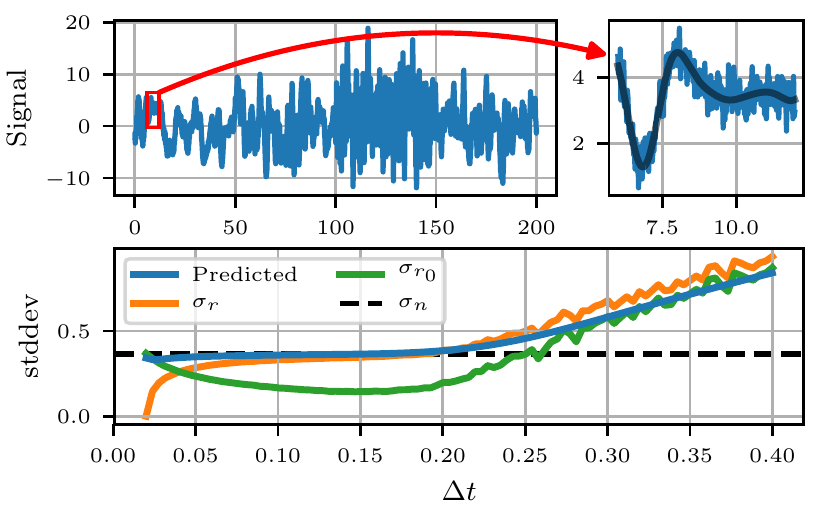}
\caption{Top: test signal and noise (right subplot is a
  detail). Bottom: standard deviations as functions of knot
  spacing. $\sigma_r$ is the empirical residual standard deviation, 
  $\sigma_n$ is the noise standard deviation, which is used in
  \cite{patron-perez15} to predict $\sigma_r$, {\it Predicted} is the \emph{Spline Error Weighting} residual
  noise prediction. $\sigma_{r_0}$ is the residual with respect to the noise-free signal $x_0(t)$.}
\label{fig:splinefit}
\end{figure}

In Figure \ref{fig:splinefit} we illustrate a simple experiment that
demonstrates the behaviour of the Spline Error Weighting residual error 
prediction \eqref{eq:residual_error}. 
In Figure \ref{fig:splinefit} top left, we show a test signal,
$x(t)$, which is the sum of a true signal, $x_0(t)$, and white
Gaussian noise $n(t)$ with variance $\sigma_n^2$. The true
signal has been generated by filtering white noise to produce a range
of different frequencies and amplitudes. In figure
\ref{fig:splinefit} top right, we show a detail of the signal, where
the added noise is visible.

We now apply a least-squares spline fit to the signal $x(t)$, to
obtain the spline $\hat{x}(t)$, defined as in \eqref{eq:spline_1d}.
This is repeated for a range of knot spacings, $\Delta t$, each
resulting in a different residual $r(t)=x(t)-\hat{x}(t)$. The residual
standard deviation $\sigma_r$ is plotted in Figure \ref{fig:splinefit}, bottom.
We make the same plot for the residual $r_0(t)=x_0(t) - \hat{x}(t)$ which measures
the error compared to the true signal. The resulting $\sigma_{r_0}$ curve has
a minimum at approximately $\Delta t=0.15$, which is thus the optimal knot
spacing. The fact that the actual residual $\sigma_r$ decreases for
knot spacings below this value thus indicates overfitting.
From this short experiment, we can also see that the implicit assumption made in
\cite{patron-perez15} that the noise standard deviation $\sigma_n$ can 
predict $\sigma_r$ is reasonable for knot spacings at or below the
optimal value. However, for larger knot spacings (at the right side of
the plot) this assumption becomes increasingly inaccurate.

\subsubsection{Selecting the knot spacing.}
\label{sec:sew_knot_spacing}
In \cite{patron-perez15} the spline knot spacing is fixed to $\Delta t=0.1$.
However, instead of deciding on a knot spacing explicitly, a more convenient
design criterion is the amount of approximation error introduced by
the spline fit. To select a suitable knot spacing,
$\Delta t$, we thus first decide on a {\it quality value}, $\hat q \in
(0, 1]$, that corresponds to the fraction of signal energy we want the
approximation to retain. For a given signal, $x(t)$, with the DFT,
${\bf X}$, we define the quality value as the ratio between the
energy, before and after spline fitting:
\begin{align}
  q(\Delta t) = \frac{\energy{{\bf H}(\Delta t) \cdot {\bf X}}}{\energy{{\bf X}}}
  \label{eq:quality}
\end{align}
To find a suitable knot spacing for the signal, we search for the
largest knot spacing $\Delta t$ for which $q(\Delta t) \geq \hat q$.

The signals ${\bf X}$ are based on the accelerometer, and gyroscope measurements,
since these contain information about both orientation and translation.
See \cite{ovren18a} for further details.

\subsubsection{Adding a robust error norm.}
In \cite{patron-perez15}, the cost function is defined as in \eqref{eq:ct_sfm_inertial},
which assumes that the measurements are drawn from a zero-mean (Gaussian) distribution.
This is a useful model for the IMU measurements, if we account for the sensor biases,
but not for the image measurements.
The image measurements are produced by tracking or feature matching over a sequence of images.
The associations made are not perfect, and the risk of producing a feature track where the measurements
do not correspond to one single 3D point is significant.
Depending on the environment, we might also have moving objects in the scene,
which can be successfully tracked, but are obviously not good landmarks.

Since such \emph{outliers} do not correspond to the geometry we are trying to estimate,
their errors can easily be orders of magnitude larger than those of the inlier set.
If the outliers are not removed, the least-squares solver will try to bring these
large errors down, even if it means that all the other measurement residuals
(those in the inlier set) are increased.
In standard structure from motion with global shutter cameras, most outliers can be removed by
enforcing geometric consistency between observed image points.
For rolling shutter cameras, enforcing geometric consistency is much harder,
because the images no longer have a single corresponding camera pose.
We instead accept that we will have at least some outliers, and try to mitigate their effect.
We do this by introducing a \emph{robust error norm} \citep{zhang97} which scales the residuals such that
large residuals have less impact.
The cost function is thus modified to its final formulation
\begin{align}
  J(\mathcal{T}, \mathcal{X}) = 
  \sum_{{\bf y}_{k, n} \in \mathcal{Y}}&
  \phi(
    {\bf y}_{k, n} - \pi({\bf T}^{-1}(t_{k, n}){\bf x}_k)
  ) \notag
  \\
  +\sum_n&||\boldsymbol{\omega}_n-\nabla_{\omega}{\bf T}(t_n)||^2_{{\bf W}_g}
  \label{eq:cvpr_cost_function}
  \\
  +\sum_l&||{\bf a}_l-\nabla^2_{a}{\bf T}(t_l)||^2_{{\bf W}_a}\,, \notag
\end{align}
where $\phi(x)$ is a robust error norm.
In \cite{ovren18a}, as well as in this work, $\phi(x)$ is the Huber norm.

\section{Rolling shutter projection}
\label{sec:projection}
In \eqref{eq:sfm_basic} and \eqref{eq:ct_sfm_basic} the landmark projection function
$\pi(\cdot)$ was defined to simply project a 3D point to its image plane location.
This formulation works fine in the case of a global shutter camera, where there is
a single camera pose for each captured image.
In a rolling shutter camera, the image rows are captured and read out sequentially,
which results in each row having its own camera pose.
This means that an image observation
\begin{align}
  {\bf y}_{k, n} = [u, v]^T = \pi({\bf T}^{-1}(t_{k, n}) {\bf x}_k)
  \label{eq:rs_projection_standard}
\end{align}
was captured at time
\begin{align}
  t_{k, n} = t^0_n + r \frac{v}{N_v} \,.
  \label{eq:rs_projection_time}
\end{align}
Here $t^0_n$ is the time of the first row of frame $n$,
$N_v$ is the number of image rows,
and $r$ is the rolling shutter \emph{image readout time}.
$r$ is simply the time it takes to read out a frame from the camera sensor.

The astute reader may have noticed a problem with equations \eqref{eq:rs_projection_standard}
and \eqref{eq:rs_projection_time}:
the projection time $t_{k,n}$ requires knowledge of the projection row, $v$,
but at the same time, the projection row also depends on the projection time!
One of the contributions of this work are to analyze different methods for solving this chicken and egg problem.
Before doing that, we will however have to replace the landmark projection function $\pi(\cdot)$.

\subsection{The rolling shutter transfer function, $\psi$}
So far we have represented a landmark $k$ as a 3D point ${\bf x}_k \in \mathbb{R}^3$.
This is, however, not the only possible parameterization.
In \cite{patron-perez15}, whose approach we follow,
a landmark is instead represented by its first observation ${\bf y}_{k, \ast}$
and a corresponding \emph{inverse depth}, $\rho_k$.

The inverse depth formulation has the nice property that it is easy to represent
points at infinity by setting $\rho_k = 0$.
It also means that the number of landmark parameters shrinks from $3N$ to $N$,
because only $\rho_k$ has to be optimized for instead of the full 3D point ${\bf x}_k$.

With the inverse depth landmark representation we redefine the image measurement process
to instead use a \emph{rolling shutter transfer function}, $\psi(\cdot)$:
\begin{align}
  {\bf y}_{k, n} &= \psi({\bf y}_{k, \ast}, {\bf T}^{-1}(t_{k,n}) {\bf T}(t_{k,\ast}), \rho_k)
  \label{eq:transfer_function}
  \\
  &= \pi \left(
  {\bf T}^{-1}(t_{k, n})
  {\bf T}(t_{k, \ast}) 
  \begin{bmatrix}
    \pi^{-1}({\bf y}_{k,\ast}) \\
    \rho_k
  \end{bmatrix}
  \right) \,. \notag
\end{align}
$\psi(\cdot)$ is called a transfer function because it transfers the reference observation ${\bf y}_{k, \ast}$,
at time $t_{k, \ast}$,
to a new measurement at image $n$, using the inverse depth, $\rho_k$, and the trajectory ${\bf T}(t)$.

For brevity, we will mostly use the shorter form $\psi(t)$, which should be understood as the projection
(reference observation transfer) at time $t$ for some landmark and trajectory.

In the following sections we describe three different strategies to implement $\psi(\cdot)$.
One important property of each method, is how well they handle the 
\emph{rolling shutter time deviation}
\begin{align}
  \epsilon(t_{k,n})=(t_{k,n}-t^0_n)\frac{N_v}{r}-\psi_v(t_{k,n})\,.
  \label{eq:rs_time_deviation}
\end{align}
This residual measures the time deviation between the requested projection time $t_{k,n}$,
and the time corresponding to the resulting image row, $\psi_v$.
We choose to express this deviation in rows (pixels), instead of time (seconds),
because this makes it easier to compare it to the reprojection error.

An ideal rolling shutter projection model should always fulfill the \emph{rolling shutter constraint}
\begin{align}
 \epsilon(t_{k, n}) = 0 \,,
 \label{eq:rs_time_constraint}
\end{align}
but we will see that relaxing this constraint can result in other benefits,
while still producing reasonable results.

\begin{figure}[tb]
\begin{center}
\includegraphics[width=\columnwidth]{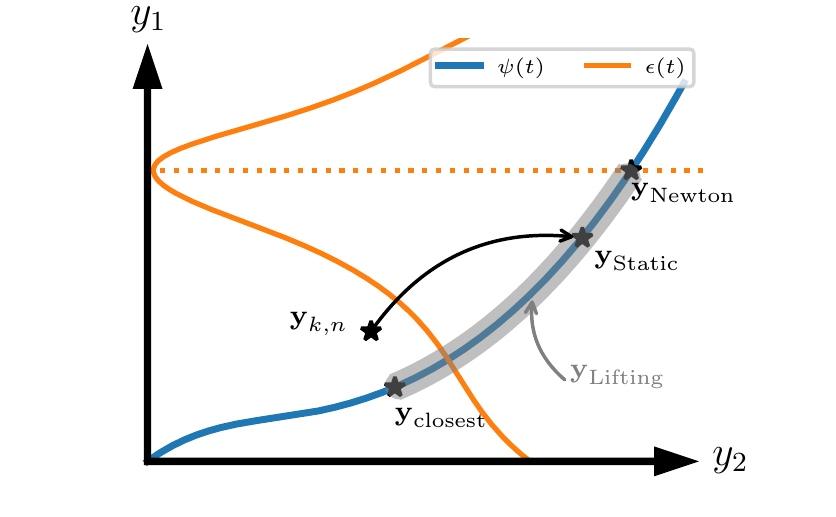}
\end{center}
\caption{Geometric illustration of the different approaches to the
  projection time problem. This is an image plane plot, where $y_1$ is
  the rolling shutter axis, ${\bf y}_{k,n}$ is the landmark observation
  in the current frame, $\psi(t)$ is the reprojection (transfer) curve for the first
  observation, as a function of spline evaluation time $t$, and $\epsilon(t)$ is the
  absolute value of the projection time deviation, plotted along the
  $y_1$ axis, as a function of $t$. In the
  illustration we can see that ${\bf y}_\text{Static}$ is obtained by
  setting the spline time to the observation time of the landmark
  ${\bf y}_{k,n}$, ${\bf y}_\text{Newton}$ is the point on the
  reprojection curve $\psi(t)$ that perfectly satisfies the projection
  time constraint $\epsilon(t)$, and ${\bf y}_\text{Lifting}$ is a
  point on $\psi(t)$ somewhere between ${\bf y}_\text{closest}$, the 
point closest to the observation, and ${\bf y}_\text{Newton}$
(depending on residual weighting).}
\label{fig:projection_time}
\end{figure}

In Figure \ref{fig:projection_time} we graphically compare the three different
methods, by plotting their possible image projections, $\psi(t_{k, n})$, together
with the time deviation $\epsilon(t_{k,n})$.

\subsection{Static projection}
One simple approach to deal with the chicken and egg problem described in
section \ref{sec:projection}, is to ignore it completely.
If we denote the observed image row by $v_{k, n}$, we set the projection time to
\begin{align}
  t_{k, n} = t^0_n + r \frac{v_{k, n}}{N_v} \,
\end{align}
and directly compute \eqref{eq:transfer_function}.

The advantage of this method is that it is fast to compute, and simple to implement.
The downside is that the projected point in general will not fulfill
the rolling shutter constraint in \eqref{eq:rs_time_constraint}.
This is shown in Figure \ref{fig:projection_time}, where the ${\bf y}_\text{Static}$ point
can end up anywhere on the $\psi(t)$ line, regardless of the value of $\epsilon(t)$.

\subsection{Newton projection}
To make sure that the rolling shutter projection time constraint in \eqref{eq:rs_time_constraint}
holds, \cite{patron-perez15} uses Newton's method to iteratively find the projection time.

To use Newton's method to solve $\epsilon(t) = 0$ we must compute $\frac{d\epsilon(t)}{dt}$,
which in turn requires computation of $\frac{d \psi(t)}{dt}$.
The transfer function $\psi(t)$ involves applying the camera projection model, $\pi(t)$,
and its inverse, $\pi^{-1}(t)$,
which means that the implementation can be quite tricky, as derivatives of these functions
are also required.
Each iteration is thus more expensive than the {\bf Static} method,
but we must also compute multiple iterations, making this a quite slow strategy.
The advantage is of course that the rolling shutter time constraint \eqref{eq:rs_time_constraint}
is now fulfilled, as we can see in Figure \ref{fig:projection_time}.

\subsection{Lifting}
The two previous methods are extremes when it comes to trading accuracy for computational efficiency.
\cite{kim16} therefore introduced a third method that aims to be more accurate than {\bf Static},
while being faster than {\bf Newton}.

This works by adding the time deviation $\epsilon(t_{k,n})$ (see \eqref{eq:rs_time_deviation})
as a new residual to the optimization problem.
The unknown projection time $t_{k, n}$ is now an additional parameter to optimize over.

The added residual makes \eqref{eq:rs_time_constraint} into a soft constraint,
which means that at best it will match the {\bf Newton} method,
and at worst give the point closest to the measured observation.
See Figure \ref{fig:projection_time} for a graphical illustration.

The described method, which we denote {\bf Lifting}, has the same
computational complexity as the {\bf Static} method.
However, since we are adding an extra residual and parameter per image observation,
the optimization problem grows larger.

\section{Spline interpolation spaces}
\label{sec:trajectories}

A time-continuous pose ${\bf T}(t)$ consists of a rotational component
${\bf R}(t)$, and a translational component ${\bf p}(t)$,
\begin{equation}
{\bf T}(t)=\begin{bmatrix}{\bf R}(t) & {\bf p}(t)\\
{\bf 0}^T & 1\end{bmatrix}\,.
\end{equation}
 Nearly all continuous camera pose representations are based on B-splines, that
define the continuous pose by blending discrete poses $\left\{{\bf
  T}_k\right\}_1^K$. In this section we introduce and compare the two trajectory
representations that are used in this work: one interpolating over
control points ${\bf T}_k\in\SE3$, and one that uses separate splines
for translation, and rotation, with control points 
${\bf p}_k\in\R3$, and ${\bf R}_k\in\SO3$, respectively.
We also analyze the theoretical difference between the two when interpolating a camera pose.

\subsection{A split spline in $\RplusSO3$}
\label{sec:split}

A regular B-spline curve in vector space $\mathbb{R}^n$ can be written:
\begin{equation}
{\bf p}(t)= \sum_{k=1}^K {\bf p}_k B(t-k\Delta t) =
\sum_{k=1}^K{\bf p}_kB_k(t)\,,
\label{eq:spline}
\end{equation}
where ${\bf p}_k\in\mathbb{R}^n$ are the
spline control points, and $B_k(\cdot)$ are the shifted B-spline basis
functions (cf. \eqref{eq:spline_1d}), that distribute the influence
of each control point in a specific time window.

Any spline of form \eqref{eq:spline} may instead be written in cumulative form:
\begin{equation}
{\bf p}(t)={\bf p}_1\tilde{B}_1(t)+\sum_{k=2}^K({\bf p}_k-{\bf p}_{k-1})\tilde{B}_k(t)\,,
\label{eq:r3_spline}
\end{equation}
where $\tilde{B}(t)$ are the corresponding {\it cumulative}
basis functions. \cite{kim95} show that this construction
is also feasible on $\SO3$, and propose to use unit
quaternions ${\bf q}_k$ as orientation control points to interpolate
\begin{equation}
{\bf q}(t)={\bf q}_1^{\tilde{B}_1(t)}\prod_{k=2}^K\exp(\log({\bf q}_{k-1}^\ast{\bf q}_k)\tilde{B}_k(t))\,.
\label{eq:so3_spline}
\end{equation}
Here ${\bf q}^*$ denotes the conjugation of the quaternion ${\bf q}$, and $\exp()$ and $\log()$
are mappings to $\text{Spin}(3)$, and its tangent space, respectively. The rationale behind
\eqref{eq:so3_spline} is the classical SLeRP interpolation \citep{shoemake85}:
\begin{equation}
{\bf q}(\lambda)={\bf q}_1\exp(\lambda\log({\bf q}_1^\ast{\bf q}_2)) \quad \lambda\in[0,1]\,.
\label{eq:slerp}
\end{equation}
The expression \eqref{eq:slerp} moves smoothly between ${\bf q}_1$
and ${\bf q}_2$ as $\lambda$ is moved from $0$ to $1$. By comparing
\eqref{eq:so3_spline} with \eqref{eq:slerp} we see that the Kim et
al.~construction is essentially a blending of SLeRP 
interpolations, within each B-spline support window. 

In summary, \cite{kim95} advocate pose interpolation with
\eqref{eq:r3_spline} for position and \eqref{eq:so3_spline} for orientation.
We will denote this as \emph{split interpolation}, or
\emph{split representation}.

\subsubsection{IMU predictions for the split interpolation.}
The IMU predictions for the split representation 
is most suitably derived using quaternion algebra,
with vectors ${\bf v} \in \R3$
embedded in pure quaternions ${\bf q_v} = \begin{pmatrix}0 & {\bf v}\end{pmatrix}^T$.
${\bf g}$ is the gravity vector, in the global coordinate frame.
We only show how to get the ideal gyroscope and IMU measurements from the trajectory,
and disregard other aspects of the IMU model, such as bias, or axis misalignment.

\begin{description}
  \item[Gyroscope prediction] 
  \begin{align}
  \begin{pmatrix}
    0\\
    \nabla_\omega {\bf T}(t)
  \end{pmatrix} =
  {\bf q}_\omega^\text{body}(t) &=
  {\bf q}^*(t) {\bf q}_\omega^\text{global}(t) {\bf q}(t)\,\text{ where} \\
  {\bf q}_\omega^\text{global}(t) &= 2 \dot{\bf q}(t) {\bf q}^*(t)
  \label{eq:split_gyroscope}
\end{align}

  \item[Accelerometer prediction]
  \begin{align}
  \begin{pmatrix}
    0\\
    \nabla_a^2 {\bf T}(t)
  \end{pmatrix} = 
  {\bf q}^*(t) 
    \begin{pmatrix}
      0\\
      \ddot{{\bf p}}(t) - {\bf g}
    \end{pmatrix}
  {\bf q}(t)
  \label{eq:split_accelerometer}
\end{align}
\end{description}

\subsection{A spline in $\SE3$}
In \cite{patron-perez15} the quaternion spline \eqref{eq:so3_spline} is
generalized to a spline construction with control points ${\bf T}_k\in\SE3$:
\begin{equation}
{\bf T}(t)=\exp(\log({\bf T}_1) \tilde{B}_1)\prod_{k=2}^K\exp(\log({\bf T}_{k-1}^{-1}{\bf T}_k)\tilde{B}_k(t))\,.
\label{eq:se3_spline}
\end{equation}
Just like in the quaternion case, this is a blend of linear interpolations on the group, within each B-spline window.

In \cite{patron-perez15} the poses to interpolate are defined as
transformations from the body frame to the global frame, \ie, 
\begin{equation}
{\bf T}({\bf R},{\bf p})=\begin{bmatrix}
{\bf R} & {\bf p}\\
{\bf 0}^T & 1\end{bmatrix}\,,
\end{equation}
where ${\bf p}$ is the spline position in the global frame, and
${\bf R}$ is the rotation from the body frame to the global
frame. Note that interpolation of ${\bf p}$ and ${\bf R}$ separately,
using \eqref{eq:r3_spline} and \eqref{eq:so3_spline} is not 
equivalent to \eqref{eq:se3_spline}. The difference 
between the two is revealed by expanding the $\SE3$  tangent,
or {\it twist} \citep{murray94}, that is used to move between two poses
in \eqref{eq:se3_spline}:
\begin{equation}
\log({\bf T}_1^{-1}{\bf T}_2)=\log\begin{bmatrix}
{\bf R}_1^T{\bf R}_2 & {\bf R}_1^T({\bf p}_2-{\bf p}_1)\\
{\bf 0}^T & 1 \end{bmatrix}\,.
\label{eq:se3_tangent}
\end{equation}
A twist $\boldsymbol{\xi}=({\bf v},\boldsymbol{\omega})\in\mathfrak{se}(3)$, consists
of a  translation ${\bf v}$ (with direction and scale), and an
axis-angle vector $\boldsymbol{\omega}$. By
exponentiating a twist times a scalar amount $\theta$ we obtain an
element in $\SE3$, with the following analytic expression:
\begin{gather}
\text{exp}(\boldsymbol{\xi}\theta)
=\text{exp}\left(\begin{bmatrix}[\boldsymbol{\omega}]_\times&
{\bf v}\\{\bf 0}^T & 0\end{bmatrix}\theta\right)=\\
\begin{bmatrix}
\text{exp}([\boldsymbol{\omega}]_\times\theta)& ({\bf
  I}-\text{exp}([\boldsymbol{\omega}]_\times\theta))[\boldsymbol{\omega}]_\times{\bf
  v}+\boldsymbol{\omega}\boldsymbol{\omega}^T{\bf v}\theta \\
{\bf 0}^T & 1
\end{bmatrix}\,,
\label{eq:twist_exponent}
\end{gather}
where $[\cdot]_\times$ is the cross product operator, \ie, $[{\bf
  a}]_\times{\bf b}={\bf a}\times {\bf b}$, see \cite[eq.~2.36]{murray94}.
In analogy with this, the twist in \eqref{eq:se3_tangent} is weighted
by a basis function value $\tilde{B}_k(t)$ and exponentiated in
\eqref{eq:se3_spline}. We can thus identify $\theta$ with $\tilde{B}_k(t)$.

\subsubsection{IMU predictions for $\SE3$.}
To compute the IMU preditions for $\SE3$, we use the same formulation as in \cite{patron-perez15}.
Here $\dot{\bf R}(t)$, $\dot{\bf p}(t)$, and $\ddot{\bf p}(t)$,
are the corresponding submatrices of $\dot{\bf T}(t)$, and $\ddot{\bf T}(t)$.
${\bf g}$ is the gravity vector, in the global coordinate frame.
Again, we only show how to get the ideal gyroscope and IMU measurements from the trajectory,
and disregard other aspects of the IMU model.

\begin{description}
  \item[Gyroscope prediction]
  \begin{align}
  \nabla_\omega {\bf T}(t) &= {\boldsymbol \omega}\, \text{ where} \\
  [{\boldsymbol \omega}]_\times &= 
  \begin{bmatrix}
    0 & -\omega_z & \omega_y \\
    \omega_z & 0 & -\omega_x \\
    -\omega_y & \omega_x & 0
  \end{bmatrix} =
  {\bf R}^T(t) \dot{\bf R}(t)
  \label{eq:se3_gyroscope}
  \end{align}

  \item[Accelerometer prediction]
   \begin{align}
  \nabla_a^2 {\bf T}(t) = {\bf R}^T(t) (\ddot{\bf p}(t) - {\bf g})
  \label{eq:se3_accelerometer}
  \end{align}
\end{description}

\subsection{Why $\SE3$ splines are problematic}
\label{sec:se3_problems}
Here we describe a number of problems with choosing $\SE3$ as the interpolation space.

\subsubsection{Translation is linked with orientation.}
\label{sec:se3_linked}
\begin{figure}[tb]
\includegraphics*[width=\columnwidth]{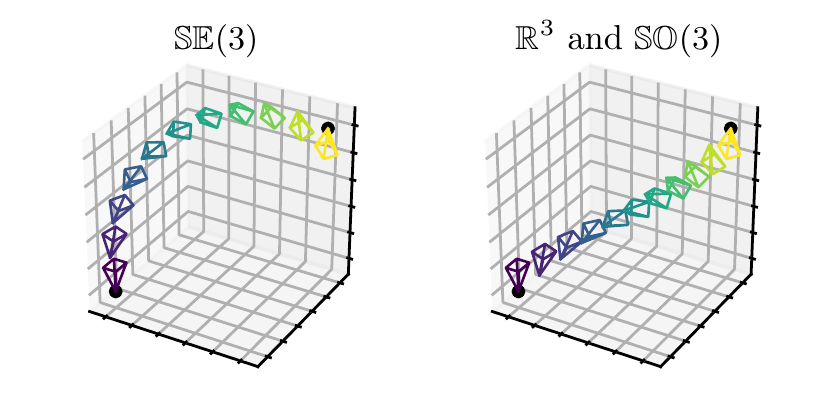}
\caption{Trajectories from interpolation of two poses on $\SE3$
  (left) and separate interpolation in $\RplusSO3$ (right). Here, start and 
  end poses differ by $150^\circ$ in orientation, which exposes the
  screw motion caused  by the $\SE3$-based interpolation.}
\label{fig:interaction}
\end{figure}
By identifying the exponentiation of \eqref{eq:se3_tangent} with \eqref{eq:twist_exponent},
when $\theta=1$, we can further identify the rotation component as
$\text{exp}([\boldsymbol{\omega}]_\times)={\bf R}_1^T{\bf
  R}_2$ (and thus $\boldsymbol{\omega}$ is parallel to the axis of rotation, which
implies $\boldsymbol{\omega}={\bf R}_1^T{\bf
  R}_2\boldsymbol{\omega}$). 
For intermediate values of $\theta$, the translation in \eqref{eq:twist_exponent}
consists of a component parallel to the rotation axis
(\ie, $\boldsymbol{\omega\omega}^T{\bf v}$) and one orthogonal to it
(\ie, $[\boldsymbol{\omega}]_\times{\bf v}$) that depends on the amount of
rotation. Unless the translation is parallel to the rotation 
axis, there will thus be an interaction between the rotation and
the translation.  The effect of this coupling of translation and
orientation is that the camera position moves along a trajectory that
spirals about the rotation axis $\boldsymbol{\omega}$, as exemplified in \figurename~\ref{fig:interaction}.
Such a motion is called a {\it screw motion} \citep{murray94}.

The implicit mechanical model in $\SE3$-based interpolation is
that the pose is manipulated by an {\it internal force and torque},
\ie, a force applied to the same fixed reference point, and with a
torque about a fixed axis in the intrinsic pose frame (such an action
is called a {\it wrench} \citep{murray94}).  
For separate interpolation of position and orientation (see section
\ref{sec:split}), pose is instead manipulated by a {\it
  generic force and torque} acting on the pose frame in different
ways at different times.

The above interpretation predicts that the $\SE3$ model would be
a good fit for \eg, cameras mounted at the end of a robot arm, and
in the idealized case also car mounted cameras \eg, dashcams.
The split interpolation model makes fewer assumptions about how the motion
changes, and is thus likely to be of more general use. 

\subsubsection{Derivative vs. body acceleration.}
\begin{figure}
  \begin{center}
    \includegraphics[width=\columnwidth]{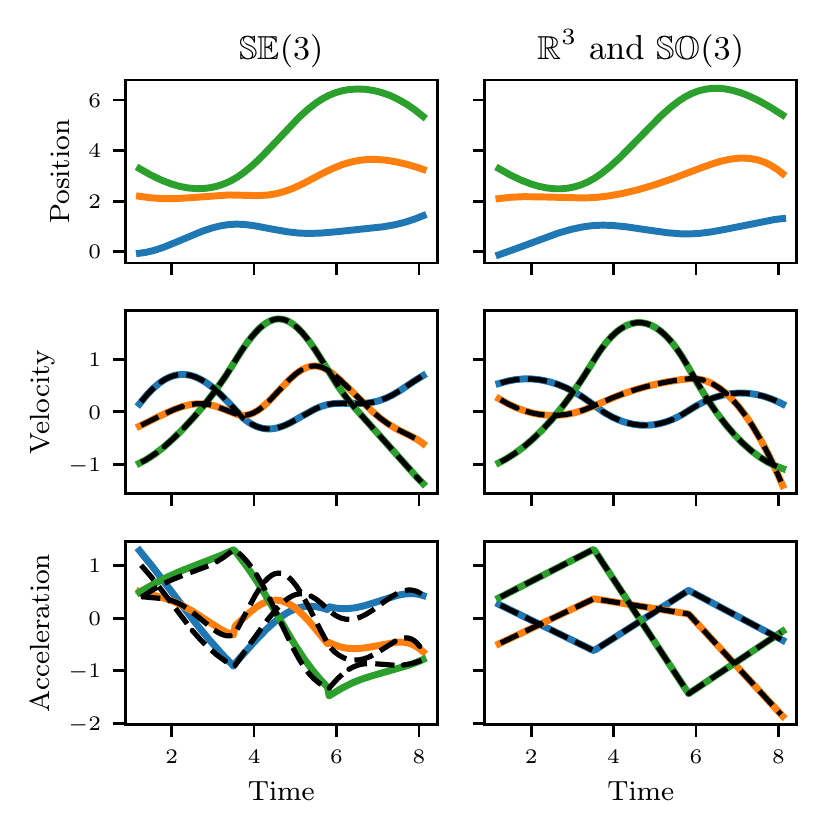}
    \caption{The problem with acceleration in $\SE3$.
    Solid and colored lines are the position, velocity, and acceleration as computed from the spline interpolation.
    Black dashed lines are the same quantities, but instead computed by numerical differentiation.}
    \label{fig:acceleration_problem}
  \end{center}
\end{figure}

To compute the accelerometer predictions, \eqref{eq:se3_accelerometer} and  \eqref{eq:split_accelerometer}, we must first compute the
linear acceleration of the body, denoted $\ddot{\bf p}(t)$.
For split interpolation this is simply the second order derivative of the $\R3$-spline, which does not impose any problem.
Using the $\SE3$ representation, $\ddot{\bf p}(t)$ is defined as a submatrix of $\ddot{\bf T}(t)$.
In section \ref{sec:se3_linked} we saw that during interpolation, $\SE3$ introduces a dependency between the translation part,
and the orientation part.
It turns out that a similar orientation-dependency problem exists also when computing the acceleration,
see \eg \cite{zefran96a}.

We illustrate this problem in Figure \ref{fig:acceleration_problem}.
Here we have constructed two pose trajectories: one in $\SE3$ and one 
split in $\RplusSO3$.
These trajectories have equal knot placements, and are designed to be as similar as possible.
For a trajectory to be well behaved, we expect that its velocity is the first order
derivative of the position (${\bf v}(t) = \frac{d {\bf p}(t)}{dt}$),
and that acceleration is the first order derivative of velocity (${\bf a}(t) = \frac{d {\bf v}(t)}{dt}$).
To test whether this holds true for the two trajectories,
we first analytically compute the position, velocity, and acceleration, 
using their respective spline formulations \citep[eqns. 4-6]{patron-perez15}.
We then compute velocity and acceleration again,
but this time using numerical differentiation of
position and velocity, respectively.
The idea is to now check whether the numerical and analytical quantities are equal.

Figure \ref{fig:acceleration_problem} clearly shows that both trajectory representations
behave as expected with respect to velocity, since the analytical and numerical results
are identical.
For acceleration, we can see that this holds true only for the split interpolation,
while $\SE3$ shows severe differences.
Only if we were to set the orientation constant (${\bf R}(t) = {\bf R}$)
does the analytical and numerical results agree, which verifies that the
problem is indeed caused by interaction with the orientation.

This means that the acceleration produced by the $\SE3$ trajectory derivative
is not the true, kinematic, body acceleration.
Accelerometer predictions computed from it, will therefore also
be inaccurate.

\subsubsection{Efficiency.}
\label{sec:se3_efficiency}
In general, evaluation of an $\SE3$ spline is
slightly more expensive, as the translation part of a spline is
evaluated using the upper right element in \eqref{eq:twist_exponent}
instead of the simpler vector difference in \eqref{eq:r3_spline}.
The $\SO3$ part is, however, the same for both methods.

Another efficiency issue has to do with evaluation of derivatives.
Here, the split $\RplusSO3$ representation allows for a potential speedup by
choosing to compute only the derivatives that are required
for each term in the visual-inertial cost function \eqref{eq:cvpr_cost_function}:
\begin{itemize}
  \item To compute the gyroscope residuals (see \eqref{eq:split_gyroscope} and \eqref{eq:se3_gyroscope}),
  only the first order orientation derivative is needed.
  However, when using $\SE3$ we must compute the full $\dot{\bf T}(t)$ matrix,
  which implicitly also calculates the superfluous linear part.
  
  \item Computing the acceleration residuals (see \eqref{eq:split_accelerometer} and \eqref{eq:se3_accelerometer})
  requires the linear acceleration, and orientation.
  In the case of split interpolation on $\RplusSO3$, the linear acceleration in $\R3$ is very efficient to compute,
  while we only need to evaluate the orientation in $\SO3$.
  In $\SE3$, we must of course compute the full $\ddot {\bf T}(t)$ matrix, which requires more computations.
\end{itemize}

\section{Experiments}
\label{sec:experiments}
In section \ref{sec:trajectories} we described two different choices of trajectory representation,
and their properties and theoretical problems.
We will now investigate what impact the identified problems have on
practical applications. In section \ref{sec:projection}, we described three different choices of rolling shutter projection methods.
We now want to see how these methods differ with respect to accuracy and runtime efficiency.
To investigate this, we perform a number of experiments on both synthetically generated, and recorded real data.

\subsection{Software}
\label{sec:kontiki}
To estimate the trajectory and 3D structure we used the open source \emph{Kontiki} framework \citep{kontiki},
which is developed by us\endnote{Kontiki will be released to the public in June 2018.}.
Kontiki is a general purpose continuous-time trajectory estimation framework,
built to be easy to extend.
Users choose a trajectory, add measurements (IMU, camera, etc.), 
and then ask the framework to find the most probable trajectory matching the measurements.
The least-squares solver uses the well known \emph{Ceres Solver} \citep{Ceres-Solver},
and for $\SE3$ calculations we use the \emph{Sophus} library \citep{libsophus}.
Kontiki is written in C++, but is mainly intended to be used with its Python frontend.

\subsection{Reconstruction method}
\label{sec:reconstruction}
All experiments follow the same reconstruction pipeline, which we describe here.

First we compute a suitable knot spacing for the splines, using the method
by \cite{ovren18a}, summarized in section \ref{sec:sew_knot_spacing}.
Since that method assumes a split spline defined on $\RplusSO3$, we get one knot spacing
for each interpolation space: $\Delta t_{\mathbb{R}^3}$ and $\Delta t_{\mathbb{SO}(3)}$.
To make the comparison with $\SE3$ fair, we set
$\Delta t = \min(\Delta t_{\mathbb{R}^3}, \Delta t_{\mathbb{SO}(3)})$,
and use this value for \emph{all} splines.
From the selected knot spacing, $\Delta t$, we then computed the 
corresponding IMU norm weights, ${\bf W}_a$ and ${\bf W}_g$,
as summarized in section \ref{sec:sew_weights}.

Like \cite{ovren18a}, we use keyframing to reduce the number of measurements,
to reduce the processing time.
In this case, we extract keyframes uniformly, spaced 10 frames apart.
We then use the adaptive non-maxima suppression method by \cite{gauglitz2011} to select the set of
landmarks and observations such that each keyframe has at most 100 observations.

Trajectories are initialized such that ${\bf p}(t) = {\bf 0}$, 
and ${\bf R}(t) = {\bf I}$, for all $t$.
Landmarks are set to points at infinity, using $\rho_k = 0$.

The robust error norm $\phi(\cdot)$ is the Huber norm, with
parameter $c=1$.

\subsection{Datasets}
\begin{figure}
\centering
\begin{subfigure}{0.49\columnwidth}
  \begin{center}
    \includegraphics[width=\columnwidth]{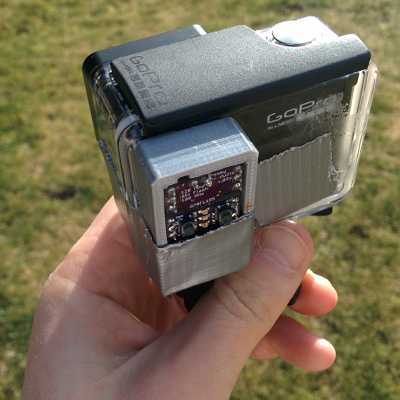}
    \caption{The GoPro camera with attached IMU logger}
    \label{fig:camera}
  \end{center}
\end{subfigure}
\begin{subfigure}{0.49\columnwidth}
  \begin{center}
    \includegraphics[width=\columnwidth]{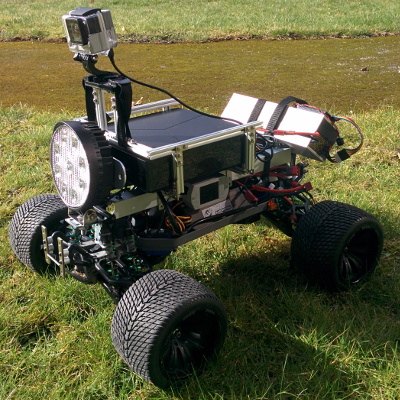}
    \caption{The radio controlled car used for the {\bf RC-Car} dataset}
    \label{fig:rccar}
  \end{center}
\end{subfigure}
\caption{Hardware used for experiments}
\label{fig:hardware}
\end{figure}
To show that the optimization behaves differently depending on the choice of interpolation space we
define the following types of motion that we want to investigate:
\begin{enumerate}
  \item {\bf Free}. Camera with free orientation. The camera orientation changes independently of the motion path.
        This simulates the case of a handheld camera, or a camera mounted on a gimbal on \eg, a UAV.
  \item {\bf Forward}. Camera locked in the forward direction
    of the path. This is similar to \eg, a dash-cam mounted in a car.
  \item {\bf Sideways}. As above but the camera is mounted looking $90^\circ$ left or right.
\end{enumerate}

Checking both the {\bf Forward locked} and {\bf Sideways locked} cases are of interest since they are known to differ
in difficulty, where the former is harder \citep{vedaldi07}.

\subsubsection{Synthetic data.}
Our synthetic data was created using the \emph{IMUSim} software package \citep{young2011}.
Since IMUSim only models IMU measurements, we implemented an extension
package\endnote{The rolling shutter extension to IMUSim can be found at \url{https://github.com/hovren/rsimusim}.}
that models rolling shutter cameras.

For each of the motion types we generated 200 random trajectories,
with matching 3D-structure, which were then processed by the simulator.
For the {\bf Forward} and {\bf Sideways} cases the ground truth trajectories
were generated using a simple motion model that tried to emulate a driving car.

The landmarks were projected into the simulated camera by finding a solution for
$\epsilon(t_{k,n}) = 0$, using the bounded root search method by \cite{brent73}.

\subsubsection{Real data.}
For the real data experiments we used two datasets\endnote{The full dataset is available from \url{http://www.cvl.isy.liu.se/research/datasets/gopro-imu-dataset/}}
called {\bf Handheld} and {\bf RC-Car}.

For both datasets, we used a \emph{GoPro Hero 3+ Black} camera,
to which we attached a custom designed IMU logger, see Figure \ref{fig:camera}.
The camera was recording using 1080p wide mode at 29.97 Hz,
while the IMU measurements were collected at 1000 Hz.
In the experiments, the raw IMU samples were resampled to 300 Hz
to reduce processing time.

\begin{figure}
  \begin{center}
    \includegraphics[width=\columnwidth]{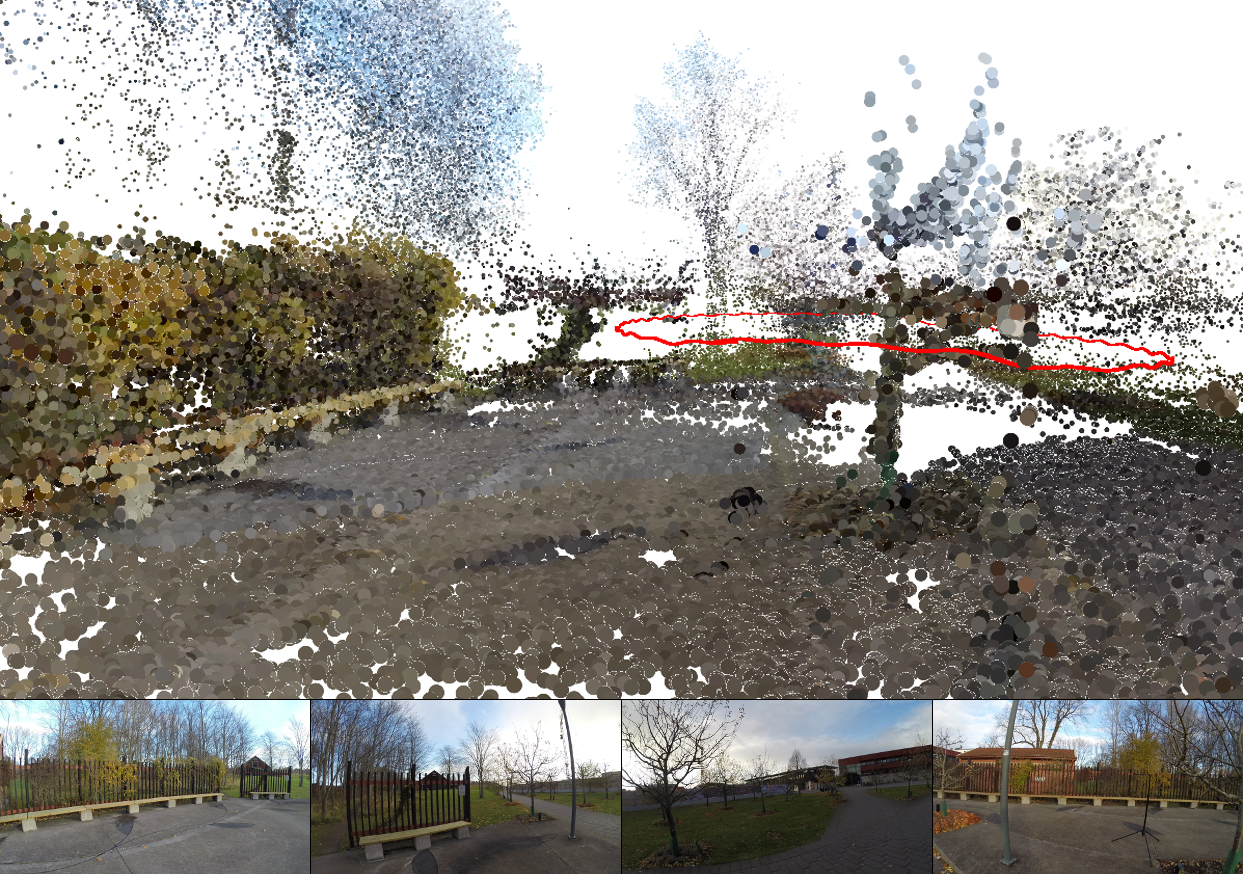}
    \caption{Rendered model estimated on the {\bf Handheld} dataset, using split interpolation. 
    Top: model rendered using Meshlab. Bottom: Sample frames from dataset.}
    \label{fig:example_handheld}
  \end{center}
\end{figure}

The {\bf Handheld} dataset was recorded while holding
the camera and walking in a loop outdoors.
Since the camera was free to rotate, it represents the {\bf Free} motion type.
Example frames from the {\bf Handheld} dataset can be found in Figure \ref{fig:example_handheld}.

In the {\bf RC-Car} dataset, the camera was attached to a radio controlled car,
see Figure \ref{fig:rccar}.
The camera was mounted pointing forwards, and thus represents the {\bf Forward}
motion type.
The RC-car was then driven in a loop over (relatively) rough terrain,
resulting in both high-frequency motion and motion blur.
Example frames from the {\bf RC-Car} dataset can be found in Figure \ref{fig:example_rccar}.

Image measurements were collected by tracking FAST features \citep{rosten10}
over subsequent frames, using the
OpenCV KLT-tracker \citep{bouguet00}.
For added robustness, we performed backtracking and discarded tracks which
did not return to within $0.5$ pixels of its starting point.
Using tracking instead of feature matching means that landmarks that
are detected more than once will be tracked multiple times by the system.

The camera-IMU extrinsics, gyroscope bias, and time offset, were 
given an initial estimate using the \emph{Crisp} \citep{ovren15} toolbox.
Since Crisp does not support accelerometer measurements,
we then refined the initial estimate using Kontiki,
described in section \ref{sec:kontiki}, by optimizing over a 
short part of the full sequence with the accelerometer bias
as a parameter to optimize.

\subsection{Trajectory representation convergence rates}
\label{sec:exp_convergence}

We want to investigate whether the choice of trajectory representation
has any impact on the reconstruction process.
By performing many reconstructions using both trajectory representations,
we can gather statistics on how the optimization cost changes over time.
Ideally we would like to compare reconstruction quality, but since
the real dataset does not have any ground truth, this is not possible.
The use of convergence rate as a metric is thus justified by the fact that it allows
us to compare the results from the synthetic and the real datasets.
Since a failed reconstruction should also cause a higher cost,
the reconstruction quality is implicitly measured by the convergence metric.

In order to gather statistics also for the real datasets (of which we have only two, longer, sequences),
we split them into a set of five seconds long, overlapping, slices, and perform the reconstructions on these instead.

In the synthetic datasets, the camera observations and IMU measurements were perturbed
by additive gaussian noise with $\sigma_{\text{image}} = 0.5$ and $\sigma_{\text{IMU}} = 0.01$,
respectively.

We always used exactly the same measurements for the $\SE3$ and split reconstructions.

\begin{figure}
  \includegraphics[width=\columnwidth]{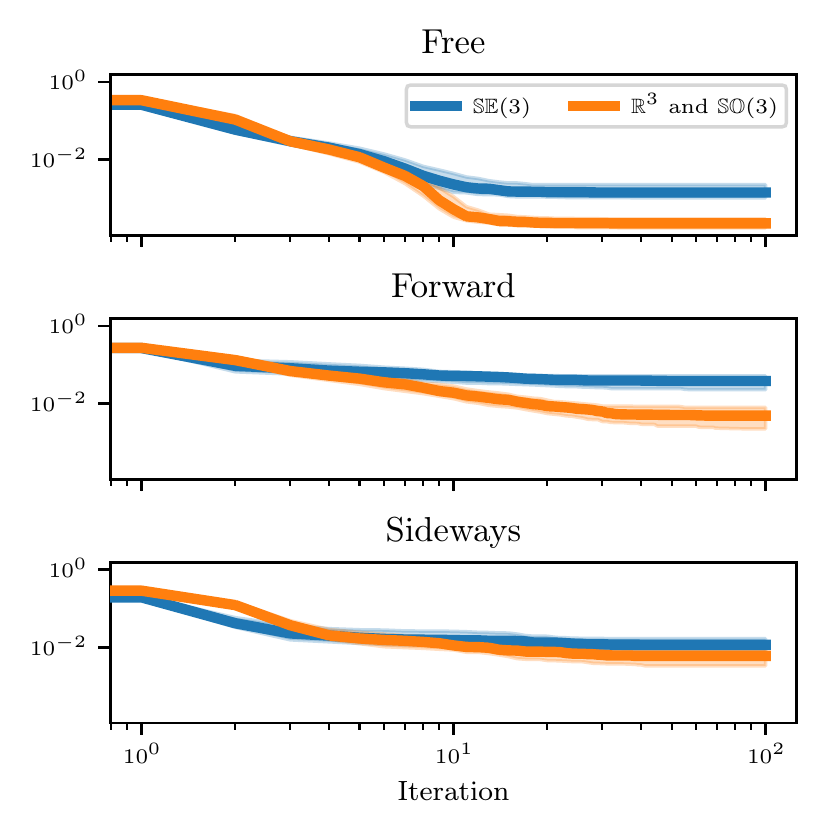}
  \caption{Convergence rate results on the synthetic dataset.
  The Y-axis shows the ratio between the current iteration cost
  and the initial cost at iteration 0.
  Solid line is the median, and the shaded area shows the distribution using the 40/60-percentiles.}
  \label{fig:convergence_synth}
\end{figure}

\begin{figure}
  \includegraphics[width=\columnwidth]{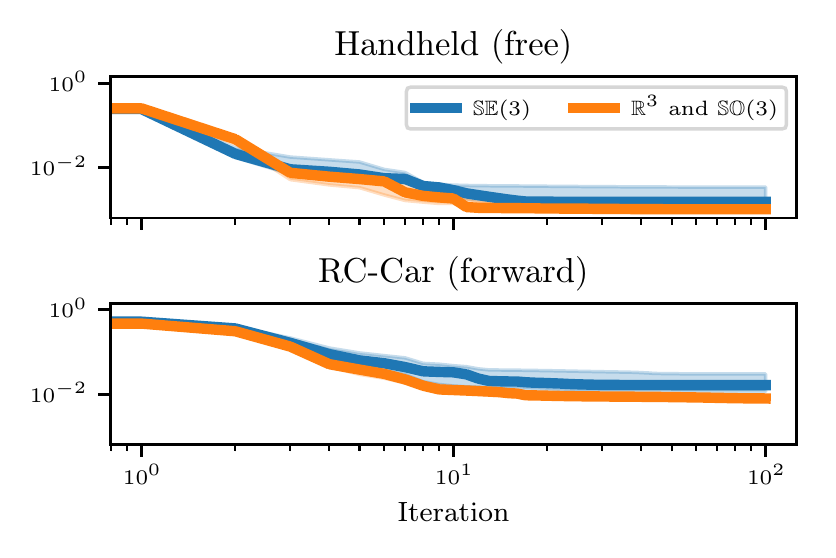}
  \caption{Convergence rate results on the real dataset.
  The Y-axis shows the ratio between the current iteration cost
  and the initial cost at iteration 0.
  Solid line is the median, and the shaded area shows the distribution using the 40/60-percentiles.}
  \label{fig:convergence_real}
\end{figure}

Figures \ref{fig:convergence_synth} and \ref{fig:convergence_real} show the
median \emph{relative cost} per iteration for the synthetic and real datasets,
respectively.
The relative cost is simply the ratio between the current iteration cost, and the initial cost at iteration 0.
To give an idea on the distribution, the shaded area shows the 40/60 percentiles of the data.
We can see that the split trajectory performs much better than $\SE3$,
giving a larger reduction in cost, which indicates a better solution.
This is true both for the synthetic and real data case, and for all motion types.

In section \ref{sec:se3_linked} we hypothesized that $\SE3$ could be a better choice for
the fixed orientation cases.
It is clear from Figure \ref{fig:convergence_synth} that the difference between
split interpolation and $\SE3$ is largest on the {\bf Free} dataset, which corroborates this.
However, $\SE3$ is clearly inferior on \emph{all} datasets, both real and synthetic,
which means that the negative aspects of $\SE3$, as described in section \ref{sec:se3_problems},
outweigh the possible benefit this might have had.

\begin{figure}
  \includegraphics[width=\columnwidth]{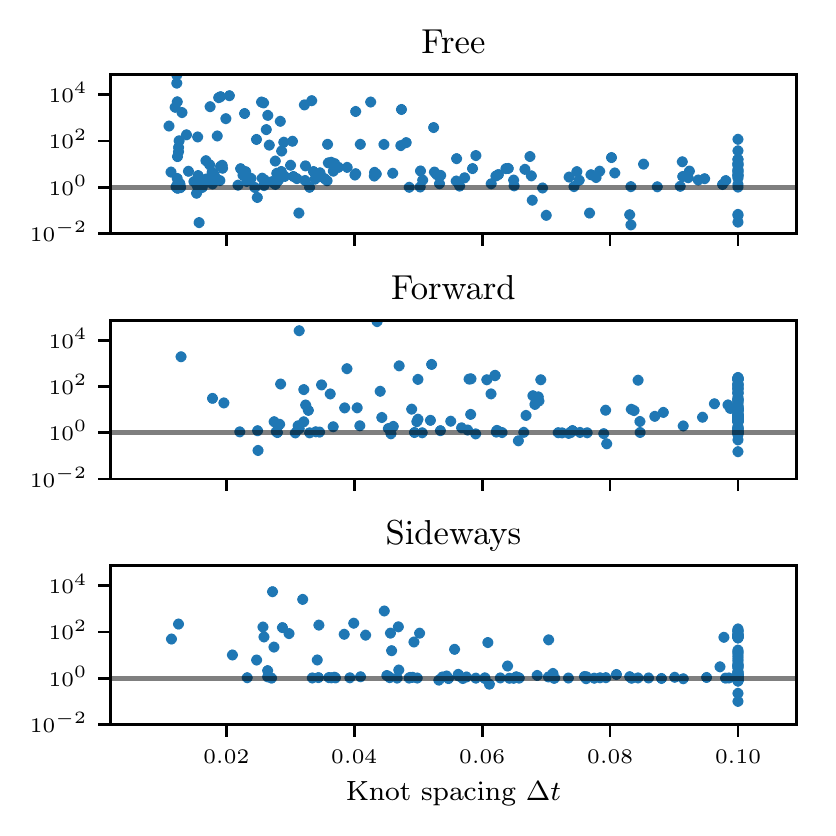}
  \caption{Distribution of relative performance between split interpolation on $\RplusSO3$ and $\SE3$ on synthetic data.
  The Y-axis shows the ratio between their respective relative costs at the final iteration.
  Samples above the line are where split representation performed better.}
  \label{fig:convergence_knotspacing_synth}
\end{figure}

\begin{figure}
  \includegraphics[width=\columnwidth]{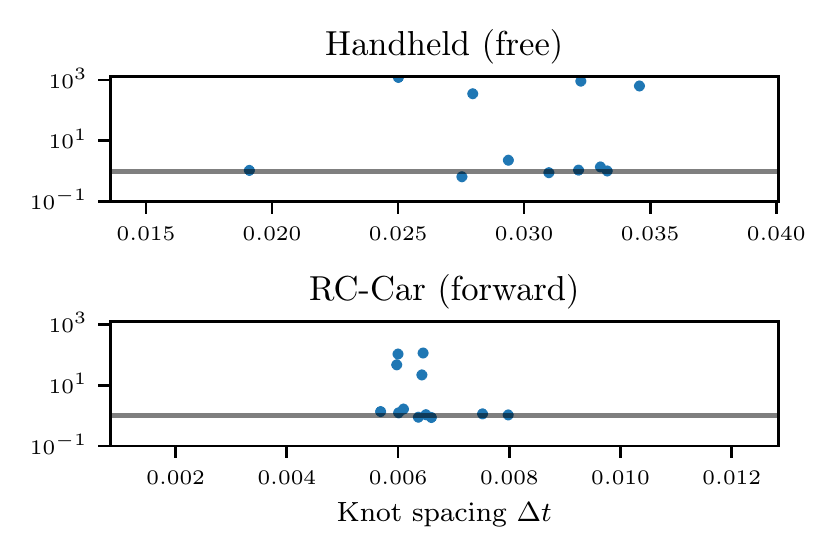}
  \caption{Distribution of relative performance between split interpolation on $\RplusSO3$ and $\SE3$ on real data.
  The Y-axis shows the ratio between their respective relative costs at the final iteration.
  Samples above the line are where split representation performed better.}
  \label{fig:convergence_knotspacing_real}
\end{figure}

To get further clues on what might affect performance, we plot the relative cost ratio for
each reconstruction as a function of the chosen knot spacing.
As we can see in figures \ref{fig:convergence_knotspacing_synth} and \ref{fig:convergence_knotspacing_real}
it is clear that $\SE3$ tends to have worse performance for small knot spacings (denser splines).

\subsection{Projection method}
\begin{figure*}[tb]
  \begin{center}
    \includegraphics[width=\textwidth]{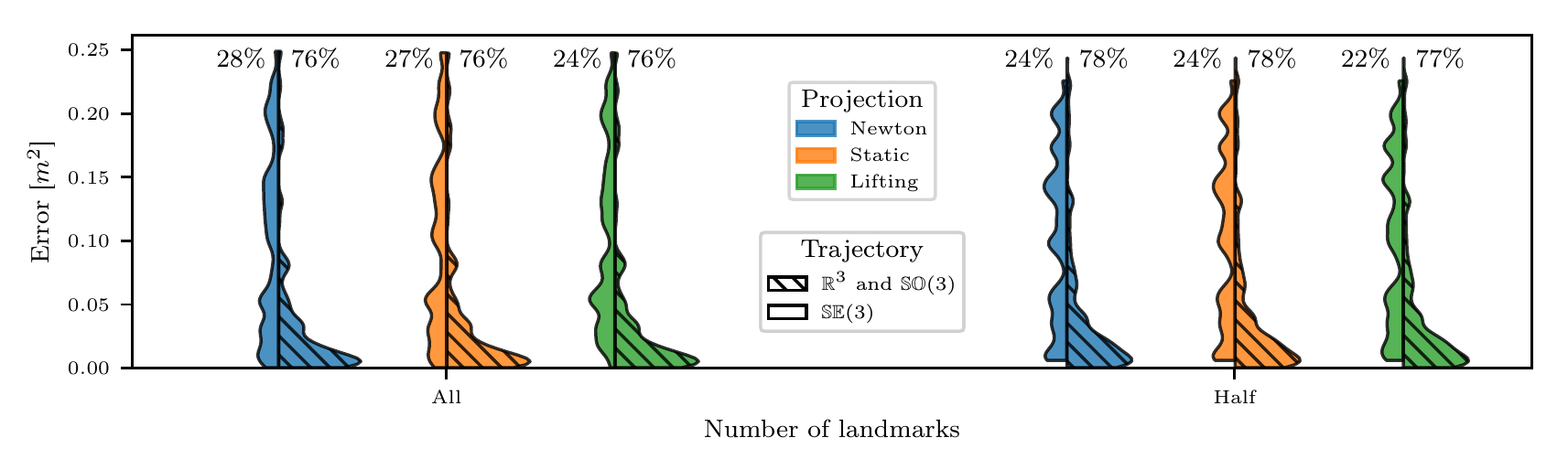}
    \caption{Distribution of errors for different combinations of trajectory representation and landmark projection method.
    The violin plots show the distribution for all errors in the inlier set, for which the error $<0.25$.
    The percentage above each violin is the inlier ratio.}
    \label{fig:projection}
  \end{center}
\end{figure*}

In section \ref{sec:projection} we described three different methods
to do rolling shutter landmark projection.
Since they differ both in implementation complexity, and runtime efficiency,
we want to make sure that slower and more complex methods actually result
in better accuracy.

In this experiment we performed reconstructions on the 200 sequences in the {\bf Free}
dataset, for all combinations of trajectory representations and projection methods.
To also investigate whether any of the methods are sensitive to the available amount of data,
we also performed the reconstructions with only half the available landmarks.

To evaluate the result we compared the estimated trajectory to the
ground truth trajectory using the soap-bubble area between the two
position trajectories, as previously suggested in \cite{hedborg12}.
The optimized trajectory was aligned to the ground truth
by using Horn's method for the translation \citep{horn87} and the orthogonal Procrustes method for
the orientation \citep{golub83}.
Since the optimization gives a trajectory with a metric scale, we do
not need to estimate any scaling factor, as was done in \cite{hedborg12}. 
For two position trajectories ${\bf f}(t)$ and ${\bf  g}(t)$, we
compute the area error numerically by trapezoid summation:
\begin{gather}
a({\bf f},{\bf g})=\sum_{k=1}^{K-1} a_\text{trap}({\bf f}(t_k),
{\bf f}(t_{k+1}),{\bf g}(t_k),{\bf g}(t_{k+1})),\text{ where}\\
a_\text{trap}({\bf a},{\bf b},{\bf c},{\bf d})=\frac{\displaystyle
  ||{\bf a}-{\bf c}||}{\displaystyle 2}(||{\bf a}-{\bf b}||+||{\bf c}-{\bf d}||)\,.
\end{gather}
This approximation of the area is only valid when sampled densely, which we do. 

In Figure \ref{fig:projection} we plot the error distributions for all tested
combinations.
Since some reconstructions fail, we choose to plot only an inlier set, which we
define as the samples with an error below $0.25\text{m}^2$.
The results in Figure \ref{fig:projection} support the conclusion
from the convergence experiment in section \ref{sec:exp_convergence}:
$\SE3$ fails more often than split interpolation, as shown by the inlier percentage.
However, even for the inlier sets, it is clear that split interpolation provides
better reconstructions since most of the distribution mass is concentrated
at lower errors.

Looking only at the results for split interpolation we can see that all three projection methods
perform more or less identically.
Also, they all benefit from more available data, which is expected.

\subsection{Efficiency}
\begin{table}
  \small\sf\centering
  \caption{Mean iteration time for different choices of interpolation space and projection method.
  The times are given relative to $\SE3$ with {\bf Newton}. Lower
  values are faster.}
  \label{tab:time_complexity}
\begin{tabular}{lccc}
\toprule
 & {\bf Newton} & {\bf Static} & {\bf Lifting}\\
\midrule
$\mathbb{SE}(3)$ & 1.00 & 0.52 & 0.54\\
$\mathbb{R}^3$ and $\mathbb{SO}(3)$ & 0.54 & 0.36 & 0.37\\
\bottomrule
\end{tabular}
\end{table}

The choice of interpolation space and reprojection method will affect the
runtime of the optimization. 
In Table~\ref{tab:time_complexity} we show the mean iteration time of
our implementations on the {\bf Free} dataset, normalized
with respect to $\SE3$ with {\bf Newton}.
Note that these timings include also the time to compute the IMU residuals.

From section \ref{sec:se3_efficiency} we hypothesized that $\SE3$ would be the slowest,
both because it is more computationally involved, and because it must 
compute superfluous derivatives for the IMU measurements.
In our implementations, split interpolation on $\RplusSO3$ is roughly twice as fast as $\SE3$ per iteration,
which supports this.

The {\bf Static} and {\bf Lifting} reprojection methods share the
same cost function, but the latter adds parameters to the optimization
which should yield a higher per-iteration cost.
The cost of the {\bf Newton} method is linear in the number of iterations taken, which is usually around 2.

Although performance is always contingent upon the specific
implementation, these practical results are consistent with the
principled discussion above.
Also, the $\SE3$ (\ie, Sophus by \cite{libsophus}) and split implementations both 
use the same {\it Eigen} (\cite{libeigen}) linear algebra library for
spline interpolation and projection computations, ensuring a fair comparison.

\subsection{Example reconstructions}
To showcase the real dataset, and to also verify that reconstruction is possible,
we performed 3D reconstruction on the original, full, sequences.
We used the pipeline from \cite{ovren18a},
which uses a split trajectory and the {\bf Static} projection method.

Since the resulting sparse reconstructions are hard to visualize, we 
densified them by triangulating new landmarks using the final trajectory.
During this densification step, the trajectory was locked, and not updated.
The trajectories and densified 3D scenes are shown in figures 
\ref{fig:example_rccar} and \ref{fig:example_handheld}.

\section{Conclusions and future work}
\label{sec:conclusions}
We have looked at two different spline-based trajectory representations,
and compared them theoretically, and experimentally.
From the presented theory we hypothesized that $\SE3$ interpolation would
perform worse than split interpolation because it makes translation
dependent on the orientation.
The experiments support this, since the $\SE3$ spline converges slower, and to a worse
result than interpolation on $\RplusSO3$, while also having a much higher failure rate.
It is also clear that $\SE3$ is less efficient,
being roughly half as fast as split interpolation.
A split $\RplusSO3$ spline also has the added flexibility of allowing splines of different densities.
Because of these findings, we recommend that researchers use a split $\RplusSO3$ spline
over an $\SE3$ spline for this type of application.

The three landmark projection methods all performed well, and produced
nearly identical results.
There was however a large difference in efficiency, with {\bf Newton} up to 
twice as slow as {\bf Lifting} and {\bf Static}.
In the context of continuous-time structure from motion, we therefore
recommend researchers to use the {\bf Static} projection method,
since it is both the fastest, and the most easy to implement.
In other applications, \eg, when the rolling shutter readout time is
also calibrated for \citep{oth13}, the difference between the methods
may be larger. Here, hybrid optimization schemes could be of interest,
where a fast method is used initially, and a more accurate one is used
in the final few iterations.

In the experiments all reconstructions were started with a trajectory
with constant position ${\bf p}(t) = {\bf 0}$ and orientation ${\bf R}(t) = {\bf I}$,
and landmarks at infinity with $\rho_k = 0$.
In contrast, discrete-time structure from motion requires a suitable initialization
to get any meaningful result.
We believe that this works because the addition of inertial measurements
gives constraints on the shape of the trajectory which can force even a bad starting state
into something useful.
From the experiments it is clear that while this initialization-free start works
quite well in general (at least for a spline defined on $\RplusSO3$), there are failure cases.
In the future we would like to investigate more robust ways to perform initialization for
visual-inertial fusion.
On the synthetic {\bf Forward} and {\bf Sideways} datasets, we have observed a correlation between
the velocity of the simulated vehicle, and the final relative cost value.
We hypothesize that the lack of a zero-velocity point makes the estimation harder,
since the integration from accelerometer measurements to velocity
assumes an initial speed of $0$.
If available, adding velocity measurements to the optimization could be
a way to remedy this.

\begin{acks}
The authors would like to thank Andreas Robinson for designing the IMU logger,
and helping out with the radio controlled car.
\end{acks}

\begin{funding}
This work was funded by the Swedish Research
Council through projects LCMM (2014-5928) and and EMC2
(2014-6227).
\end{funding}

\theendnotes

\bibliographystyle{SageH}
\bibliography{local.bib}

\end{document}